\documentclass{article}

\usepackage{microtype}
\usepackage{graphicx}
\usepackage{subfigure}
\usepackage{booktabs} 
\usepackage{tabularx}

\usepackage{hyperref}

\newcolumntype{Y}{>{\centering\arraybackslash}X}

\usepackage{algpseudocode}
\usepackage[accepted]{icml2022}

\usepackage{amsmath,amsfonts,bm}

\def\eqref#1{equation~\ref{#1}}

\def\1{\bm{1}}

\def\vepsilon{{\bm{\varepsilon}}}
\def\va{{\bm{a}}}

\def\vg{{\bm{g}}}

\def\vw{{\bm{w}}}
\def\vx{{\bm{x}}}
\def\vy{{\bm{y}}}
\def\vz{{\bm{z}}}

\def\mA{{\bm{A}}}
\def\mB{{\bm{B}}}

\def\mI{{\bm{I}}}

\def\mL{{\bm{L}}}

\def\mV{{\bm{V}}}
\def\mW{{\bm{W}}}

\DeclareMathAlphabet{\mathsfit}{\encodingdefault}{\sfdefault}{m}{sl}
\SetMathAlphabet{\mathsfit}{bold}{\encodingdefault}{\sfdefault}{bx}{n}

\def\gG{{\mathcal{G}}}

\def\gN{{\mathcal{N}}}

\def\gS{{\mathcal{S}}}
\def\gT{{\mathcal{T}}}

\def\gX{{\mathcal{X}}}
\def\gY{{\mathcal{Y}}}

\newcommand{\E}{\mathbb{E}}

\newcommand{\R}{\mathbb{R}}

\newcommand{\KL}{D_{\mathrm{KL}}}

\DeclareMathOperator{\diag}{diag}

\newcommand*{\inner}[2]{\left\langle#1,#2\right\rangle}
\newcommand*{\norm}[1]{\left\|#1\right\|}
\newcommand*{\card}[1]{\left|#1\right|}

\newcommand*{\prob}[1]{\mathbb{P}}
\usepackage{todonotes}

\def\srec{\text{S-REC}}

\def\gconv{\circledast_{G}}
\usepackage{amsthm}
\newtheorem{theorem}{Theorem}[section]

\newtheorem{lemma}[theorem]{Lemma}
\theoremstyle{definition}
\newtheorem{definition}[theorem]{Definition}

\theoremstyle{remark}

\usepackage{multirow}
\usepackage{url}
\newcommand\numberthis{\addtocounter{equation}{1}\tag{\theequation}}

\newcommand{\STAB}[1]{\begin{tabular}{@{}c@{}}#1\end{tabular}}

\usepackage{amssymb}
\usepackage{graphicx}
\usepackage{caption}
\usepackage{wrapfig}
\usepackage{adjustbox}
\usepackage{enumitem}
\usepackage[utf8]{inputenc} 
\usepackage[T1]{fontenc}    
\usepackage{url}            
\usepackage{amsfonts}       
\usepackage{nicefrac}       
\usepackage{xcolor}         
\usepackage{multirow}

\icmltitlerunning{Equivariant Priors for Compressed Sensing with Unknown Orientation}

\begin{document}

\twocolumn[
\icmltitle{Equivariant Priors for Compressed Sensing with Unknown Orientation}



\icmlsetsymbol{equal}{*}
\icmlsetsymbol{intern}{$\dagger$}

\begin{icmlauthorlist}
\icmlauthor{Anna Kuzina}{intern,vu} 
\icmlauthor{Kumar Pratik}{equal,qu}
\icmlauthor{Fabio Valerio Massoli}{equal,qu}
\icmlauthor{Arash Behboodi}{qu}
\end{icmlauthorlist}

\icmlaffiliation{vu}{Vrije Universiteit Amsterdam, Netherlands. \, $\dagger$ Work done during the internship in Qualcomm AI research.}
\icmlaffiliation{qu}{Qualcomm AI Research, Qualcomm Technologies Netherlands B.V. Qualcomm AI Research is an initiative of Qualcomm Technologies, Inc.}

\icmlcorrespondingauthor{Anna Kuzina}{av.kuzina@yandex.ru}

\icmlkeywords{Machine Learning, ICML}

\vskip 0.3in
]



\printAffiliationsAndNotice{\icmlEqualContribution} 

\begin{abstract}


In compressed sensing, the goal is to reconstruct the signal from an underdetermined system of linear measurements. Thus, prior knowledge about the signal of interest and its structure is required. Additionally, in many scenarios, the signal has an unknown orientation prior to measurements. To address such recovery problems, we propose using equivariant generative models as a prior, which encapsulate orientation information in their latent space. Thereby, we show that signals with unknown orientations can be recovered with iterative gradient descent on the latent space of these models and provide additional theoretical recovery guarantees. We construct an equivariant variational autoencoder and use the decoder as generative prior for compressed sensing. We discuss additional potential gains of the proposed approach in terms of convergence and latency. 
\end{abstract}


\section{Introduction}
Compressed sensing (CS) deals with the problem of reconstructing an unknown underlying signal $\vx \in \mathbb{R}^m$ from $m$ linear measurements generated as
\begin{equation} \label{eq:cs_main}
    \vy = \mathbf{A}\vx + \vepsilon
\end{equation}
where $\mathbf{A} \in \mathbb{R}^{m\times n}$ is a task-specific measurement matrix and $\vepsilon$ is additive noise. A wide range of practical problems can be formulated as a recovery of an unknown signal from the noisy linear measurements.
Medical applications include computed tomography \citep{chen2008prior}, where one seeks to recover an unknown image from a sinogram (set of linear measurements) and rapid MRI \citep{lustig2007sparse}, where an MR image is reconstructed from an undersampled k-space. It is also used in wireless communication for channel estimation problems \citep{paredes2007ultra}. Single-pixel imaging \citep{duarte2008single} is another example of linear inverse problems.

Since the number of measurements is assumed to be smaller than the signal space dimension, this system of linear equations is under-determined, and extra assumptions are needed to recover the solution. One common approach would be to use the sparsity assumption about the signal of interest in a given basis \citep{tibshirani1996regression,candes_stable_2006,donoho_compressed_2006}. There are other notions of structure beyond sparsity such as low rankness. More complicated structures can be captured using generative modelling. This field is usually referred to as generative compressed sensing. Generative CS  uses a pre-trained generative model to capture the prior over the space of signals. In this case, the signal recovery is done by finding the latent code of the generative model, for example, using gradient descent \cite{Bora2017-as}. There are, however, some challenges in using generative priors for compressed sensing, such as convergence and latency issues. Gradient descent methods sometimes need many iterations and restarts for convergence (see \cite{Whang2021-if} and references and discussions therein).

In this work, we consider a more general setup, where along with the known linear forward operator, the input is also transformed by an unknown group operation. A typical example is a rotation by an angle $\alpha$. Such examples occur in many imaging applications where the signal pose is unknown before the imaging. Another example is cryo-electron microscopy \cite{singer_mathematics_2018}. We are interested in this problem: 
\begin{equation}
    \vy = \mathbf{A} \gT_g \vx  + \vepsilon,
\end{equation}
where $\gT_g: \gG \times \gX \rightarrow \gX $ is an action of a group element $g$ in the group $\gG$ on the signal space $\gX$. For example, it can be a rotation by some angle which is, in many cases, captured by a linear transformation. The group transformation $\gT_g$ is assumed to be unknown.

To use standard generative priors for this task, the signal pose should be estimated jointly with the latent code. It is therefore natural to include group action information in the latent space. In this paper, we propose equivariant generative priors for this task. A generative model $G(\cdot)$ is equivariant if for group representations $\gT_g^x$ on the signal space and $\gT_g^z$ on the latent space, we have $\gT_g^xG(\vz)=G(\gT_g^z \vz)$ for all $g\in\gG$ and $\vz$. In this way, unknown transformations are coded in the latent space and the task boils down to the estimation of the transformed (e.g. rotated) latent code.

In this work we focus on the group of rotations, however, our approach can be also applied to other groups. 
We show how the current generative compressed sensing approaches can be adapted to the setup of the unknown signal orientation. We compare different baseline generative priors in this setup and propose a rotation equivariant prior based on equivariant variational autoencoder model, which can equally be used for a conventional compressed sensing problem and, at the same time, is well equipped to deal with the unknown orientation of the concerned signal. 

The contributions of this paper are as follows. We propose using equivariant generative priors for solving compressed sensing problems with unknown rotations. We show how standard recovery guarantees for generative priors are transferred to the equivariant models and compressed sensing with unknown orientation. We build an equivariant variational autoencoder. We experimentally show that the proposed prior is useful for both the conventional and the rotation-aware compressed sensing. In particular, these models provide promising gains in terms of latency and convergence. In addition, integrating the inductive bias about rotation equivariance into the model leads to a more parameter efficient generative priors.

\section{Related Work} \label{sec:related_works}

\paragraph{Generative Compressed Sensing} \label{sec:related_gen_cs}
It is common to address a compressed sensing problem using the maximum likelihood approach. Assuming that the additive noise $\vepsilon$ in the linear inverse problem (\ref{eq:cs_main}) is standard Gaussian, the likelihood would also follow Gaussian distribution $p(\vy|\vx) = \mathcal{N}(\vy| \mA\vx, \mI)$. Maximum-Likelihood estimation of the unknown signal is, therefore, a solution of the following optimization problem:
\begin{equation}\label{cs:mle}
    \vx^* = \arg\min_\vx \| \vy - \mathbf{A}\vx\|_2^2.
\end{equation}
\cite{Bora2017-as} proposed to use a generative model as a prior in this setup. It is assumed that the generative model is equipped with the generator $G(\cdot)$, which maps latent variables $\vz$ (often assumed to be standard Gaussian) to the space of signals. This way, generative model serves as a regularizer and the optimization problem (\ref{cs:mle}) can be reformulated as
\begin{align}
    \vx^* &= G(\vz^*),\\
    \vz^* &= \arg \min_{\vz} \|\vy - \mathbf{A} G(\vz)\|_2^2. \label{eq:cs_mle}
\end{align}
Using MLP-VAE and DCGAN models as a prior was shown to perform better than the sparsity prior \cite{Bora2017-as}. There are many follow-up works on better optimization over latent space \cite{daras_intermediate_2021,lei_inverting_2019}. The authors in \cite{Hussein2020-yk} introduce the Image-Adaptive GAN model, where the generator is updated together with the latent code. Normalizing Flow model is used as prior in \cite{Asim2019-lv} and shown that it outperforms previous GAN-based approaches for a variety of inverse problems, including compressed sensing. 

It was observed that the maximum likelihood estimation (\eqref{eq:cs_mle}) usually requires a proper regularization for faster convergence. Natural way to impose such regularization would be maximum a posteriori estimate of the unknown signal:
\begin{equation}
    \vx^{\text{MAP}} = \arg \max_\vx \log p(\vy|\vx) + \log p(\vx).
\end{equation}

MAP formulation was introduced in \cite{Whang2021-fj} and can be as well written in terms of the generative prior and Gaussian noise model:
\begin{align}
    &\vx^{\text{MAP}} = G(\vz^*),\\
    &\vz^* = \arg \min_\vz \|\vy - \mathbf{A} G(\vz)\|_2^2 + \log p_{G}(G(\vz)). \label{eq:cs_map}
\end{align}
It is worth mentioning that the latter term of \eqref{eq:cs_map} limits the applicability of the formulations to the generative priors with the explicit density functions, e.g. normalizing flows.  

In some applications it is also beneficial to recover the whole distribution of the unknown signal $\vx$ given the observation $\vy$. Since the true posterior distribution is either too hard or even impossible to recover, one can use stochastic variational inference (SVI) to get the approximate posterior:
\begin{equation}
    q_\vx^* = \arg\min_{q \in \mathcal{Q}} \KL\left[q(\vx) \| p(\vx|\vy) \right].
\end{equation}
This approach was recently applied to compressed sensing \cite{Whang2021-ro}. It was proposed to construct $q_\vx$ in such a way that it comprises of two normalizing flow models. The first, pre-generator $q_\vz$, is attached to a prior generative model and is trained to match the posterior distribution of the unknown signal. 
\begin{equation}
    q_\vz^* = \arg\min_{q_\vz} \KL\left[ q_\vz\|p_\vz\right] + 
    \E_{q_\vz}\left[\|\vy - \mathbf{A}G(\vz)\|_2^2 \right].
\end{equation}
The problem can be further amortized \citep{kingma2013auto, rezende2014stochastic}, so that the variational posterior is conditioned on the observation $q_{\vx}(\vx|\vy)$. This significantly reduced the computational complexity of the method. The point estimate of the signal can also be obtained in this case. One can sample from the variational posterior and average the results, which gives a Monte-Carlo estimate of the conditional expectation $\E\left[\vx|\vy\right]$.

\paragraph{Equivariant Generative Models}
Group equivariant neural networks \citep{cohen2016group} incorporate symmetries of the data in the architecture. The authors in \cite{e2cnn} provide a framework to construct neural networks equivariant under all the isometries of  $\mathbb{R}^2$ plane. Reincorporating symmetries in the architecture was shown to improve generalization and sample efficiency for discriminative models. In \cite{Karras2021-yx} authors show that rotation and translation equivariance can be beneficial for GANs as well.  In this work, we introduce VAE with fully equivariant latent space. Benefits of the equivariant decoder were also shown in \cite{Bepler2019-qi}, where authors use a spacial generator network, which allows representing rotation and translation as a coordinate space transformation. Homeomorphic-VAE \cite{falorsi2018explorations} focuses on generative models with latent space that has SO(3), in general, a Lie group, values. In this work, we consider a conventional Euclidean latent space with real-valued entries. On this space, we deﬁne the action of a discrete group of 2D rotations ($C_n$). 

Another promising direction in this area is employing equivariant normalizing flow models. Recent works have introduced equivariant generative models based on continuous normalizing flows  \cite{kohler2020equivariant, satorras2021n}, which are, however difficult to scale. 

\paragraph{Compressed Sensing and Unknown Orientation}
In this work, we focus on the reconstruction of the signal with an unknown orientation. The authors in \cite{davenport2007smashed} consider reconstructing the rotation angle of the input and focuses primarily on compressive classiﬁcation and not signal reconstruction. We aim at restoring the signal regardless of the fact that it was rotated or not. Multi-reference alignment \cite{bendory2017bispectrum} aims at solving inverse problems from multiple observations of a single source in different unknown orientations. The main challenge, and focus of research, in these works, is the alignment of these observations. For us, there is only a single observation, and the alignment step does not apply. To the best of our knowledge, this is the first work considering compressed sensing problems with unknown orientation.

\section{Methodology}

\subsection{Background: Variational Autoencoder}\label{sec:vae}
Variational Autoencoder (VAE) \citep{kingma2013auto, rezende2014stochastic} is a deep generative model which models a joint distribution of observed random variables $\vx \in \mathcal{X}$ (e.g. $\mathcal{X} = \mathbb{R}^n$) and latent variables $\vz \in \mathbb{R}^k$ as $p_{\theta}(\vx, \vz) = p_{\theta}(\vx|\vz)p(\vz)$. The model is trained to maximize the marginal likelihood $p_{\theta}(\vx)$ for a given set of points $\vx_1, \dots \vx_N$. Amortized version of variational inference \citep{jordan1999introduction} is used to obtain a tractable objective: by introducing a variational posterior $q_{\phi}(\vz|\vx)$, also refered to as encoder, one can obtain a lower bound on the intractable marginal likelihood (ELBO):
\begin{align*}
    \mathcal{L}(\theta, \phi) = \E_{p_e(\vx)}\large[&\E_{q_{\phi}(\vz|\vx)} \ln p_{\theta}(\vx|\vz) \\ 
    &\quad-\KL(q_{\phi}(\vz|\vx) \| p(\vz)) \large]. \numberthis \label{eq:elbo}
\end{align*}

\subsection{Equivariant VAE}\label{sec:eq_vae}
\paragraph{Requirements for the Equivariant Latent Space} We start by discussing the requirements, which encoder and decoder of the VAE should satisfy. Forward pass through the VAE model consists of three steps. First, we apply a neural network parameterized by $\phi$ to get parameters of the variational posterior $q_{\phi}(\vz|\vx)$. Secondly, we sample from this distribution $\tilde{\vz} \sim q_{\phi}(\vz|\vx)$. And, on the third step, we push $\tilde{\vz}$ through the neural network parameterized by $\theta$ which returns parameters of the generative distribution $p_{\theta}(\vx|\tilde{\vz})$. 

Consider a group $\gG$. Our goal is to construct VAE, in which latent space is equivariant under the action of the group element $g \in \gG$. Let $\gT_g^x: \gG \times \mathcal{X} \rightarrow \mathcal{X}$ be a transformation of a data point under the group action $g$ and $\gT_g^z: \gG \times \mathbb{R}^k \rightarrow \mathbb{R}^k$ be a transformation of a latent code under the same action $g$. Then, for the encoder to be equivariant, we aim for the following property to hold:

\begin{align}
    &\gT_g^z \vz \stackrel{d}{=} \vz_g, \text{where}\label{eq:eq_encoder}\\ 
     &\vz_g \sim q_{\phi}(\vz| \gT_g^x \vx),\\
     &\vz \sim q_{\phi}(\vz| \vx).
\end{align}
where $\stackrel{d}{=}$ stands for equality in distribution. In other words, sampling latent code for variational posterior conditioned on the transformed input should be the same as transforming a latent code which is sampled from the variational distribution conditioned on the non-transformed input.  

Furthermore, symmetric property should hold for the decoder network. Namely,
\begin{align}
    &\gT_g^x \vx \stackrel{d}{=} \vx_g, \text{where} \label{eq:eq_decoder}\\
     &\vx_g \sim p_{\theta}(\vx| \gT_g^z \vz),\\
     &\vx \sim p_{\theta}(\vx| \vz).
\end{align}

We now discuss how to construct the Encoder and Decoder networks to ensure that both requirements (\ref{eq:eq_encoder}) and (\ref{eq:eq_decoder}) hold.

\paragraph{Equivariant Encoder}
The common choice of the variational posterior distribution $q_{\phi}(\vz|\vx)$ is Gaussian 
\begin{align*}
    q_{\phi}(\vz|\vx) &= \mathcal{N}(\vz| \mu, \Sigma), \numberthis \label{eq:var_posterior}\\
    \mu &= \mu_{\phi}(\vx),\\
    \Sigma &= \mL\mL^T, \, \mL = L_{\phi}(\vx).
\end{align*}

In general case $\mL$ is a lower triangular matrix, which can be obtained by Cholesky decomposition of $\Sigma$. In practice, the covariance matrix is often chosen to be diagonal. That is $\mL$ is also diagonal: $L_{\phi}(\vx) = \diag \sigma_{\phi}(\vx)$. 

Such parametrization is practical for several reasons. Firstly, Gaussian distribution is reparametrizable:
\begin{equation}
    \vz = \mu +  \mL\varepsilon,
\end{equation}
This is crucial for backpropogation through the ELBO (\ref{eq:elbo}). Secondly, the KL-term between variational posterior and prior is to be computed at each iteration. The prior is usually chosen to be standard Gaussian, which gives a closed form for the KL-divergence:
\begin{align*}
    \KL\left[q_{\phi}(\vz|\vx)\|p(\vz)\right] = \tfrac{1}{2}\large[-&\log\det\Sigma - d + \\
    &\text{tr}\Sigma + \mu^T\mu \large]. \numberthis \label{eq:gaus_kl}
\end{align*}
With the diagonal covariance matrix computing the log-determinant in \eqref{eq:gaus_kl} reduces to a summation of $d$ scalars.

In this work, we also focus on the Gaussian $q_{\phi}$. Then it is enough to match the first and second moments of $\gT_g^z \vz$ and $\vz_g$ to satisfy the desired properties of the equivariant latent space. 
When it comes to $\vz_g$, it follows a Gaussian distribution, which gives us the following moments: 
\begin{align}
    &\E \vz_g = \mu_{\phi}(\gT_g^x \vx), \\
    &\E(\vz_g - \E \vz_g) (\vz_g - \E \vz_g)^T = L_{\phi}(\gT_g^x \vx)L_{\phi}(\gT_g^x \vx)^T.
\end{align}

Next, we would like to find the moments of $\gT_g^z \vz$, knowing that $\vz$ is a Gaussian random variable (\ref{eq:var_posterior}). The mean value is given by
\begin{equation}
    \E \gT_g^z \vz = \E \gT_g^z[\mu + \mL\varepsilon] =  \gT_g^z[\mu].
\end{equation}
And for the second moment, we have:
\begin{align}
    \E &(\gT_g^z \vz - \E  \gT_g^z \vz)(\gT_g^z \vz - \E  \gT_g^z \vz )^T \\
    &=\E (\gT_g^z[\mu + \mL\varepsilon] + \gT_g^z \mu)(\gT_g^z[\mu + \mL\varepsilon] + \gT_g^z\mu)^T \\
    &=\E \gT_g^z[\mL\varepsilon]\gT_g^z[\mL\varepsilon]^T \\
    &=\gT_g^z[\mL]\gT_g^z[\mL]^T.
\end{align}

This gives us the following condition on the mean function:
\begin{equation}\label{eq:eq_encoder_mean_final}
     \mu_{\phi}(\gT_g^x \vx) = \gT_g^z[\mu_{\phi}(\vx)].
\end{equation}

This property can be satisfied when using an equivariant neural network to model the mean function. The covariance, on the other hand, should satisfy:
\begin{equation}\label{eq:eq_encoder_var_final}
    L_{\phi}(\gT_g^x \vx)L_{\phi}^T(\gT_g^x \vx) = \mathcal{T}_g^z[L_{\phi}(\vx)] \left(\gT_g^z[L_{\phi}(\vx)]\right)^T.
\end{equation}

The main restriction that results from the (\ref{eq:eq_encoder_var_final}) is that the diagonal covariance matrix can violate the desired equivariance property. As mentioned earlier, one usually chooses $L_{\phi}(\vx) = \diag \sigma(\vx)$. In this case $L_{\phi}(\gT_g^x \vx)$ will always be diagonal. However, $\mathcal{T}_g^z[L_{\phi}(\vx)]$ is not necessarily staying diagonal. Therefore, we have to use a non-diagonal covariance matrix. We still need to be able to sample $\vz$, compute log-determinant on the forward pass and ensure that the matrix is positive definite. To this end, we propose to use a full-rank covariance matrix in our model:
\begin{equation}
    \Sigma = V_{\phi}(\vx)V_{\phi}(\vx)^T + \eta \mI,
\end{equation}

where $V_{\phi}(\cdot)$ is an equivariant function, which outputs a full-rank matrix.  We ensure that the resulting covariance matrix is positive definite by adding a small positive perm to the diagonal elements.

\paragraph{Equivariant Decoder}
For the decoder to be equivariant, we should ensure a similar property as we did for the encoder. We consider two distributions to model the conditional likelihood $p_{\theta}(\vx|\vz)$. For RGB images, we can use Gaussian distribution. The same assumptions about the mean function and covariance should be satisfied in this case. 

For grey-scale images, Bernoulli distribution is used: $p_{\theta}(\vx|\vz) = \mathcal{B}e(\vx | f_{\theta}(\vz))$. In this case, it is straightforward to respect the property (\ref{eq:eq_decoder}): 
\begin{equation}
f_{\theta}(\gT_g^z \vz)  = \mathcal{T}_g^x[f_{\theta}(\vz)],
\end{equation}

which means that we should just ensure that $f_{\theta}$ is an equivariant function. We summarize the forward pass through the resulting model in the Algorithm \ref{alg:vae_forward}.

\begin{algorithm}[t]
	\caption{Forward pass through equivariant VAE}
	\label{alg:vae_forward}
	\begin{algorithmic}
	    \State \hskip-3mm \textbf{Input}: $x$
	    \State $\mu = \mu_{\phi}(\vx)$  \Comment{Forward pass through the encoder}
	    \State $\mV = V_{\phi}(\vx)$
		\State $\Sigma = \mV\mV^T + \eta \mI$
		\State $L = \text{Chol}(\Sigma)$ \Comment{Cholesky decomposition of $\Sigma$}
		\State $\widetilde{\vz} = \mu + \mL \varepsilon, \, \varepsilon \sim \mathcal{N}(0, I)$
		\State $\text{KL} = \tfrac12\sum_i\left(-\log \mL_{ii} - 1 + \mL_{ii}^2 + \mu_i^2 \right)$
		\State $\mathrm{p}_x = f_{\theta}(\widetilde{\vz})$ \Comment{Forward pass through the decoder}
		\State $\text{Re} = -\log p_{\theta}(\vx| \mathrm{p}_\vx)$ \Comment{Reconstruction Loss}
		\State $\mathcal{L}(\phi, \theta) =  - \text{Re} - \text{KL}$ \Comment{Compute ELBO}
		\State  \hskip-3mm \textbf{Return}: $ - \mathcal{L}(\phi, \theta)$
	\end{algorithmic}
\end{algorithm}

\subsection{Generative Compressed Sensing with Unknown Orientation} \label{sec:gen_cs_rotation}

We consider the following forward process:
\begin{equation}
    \vy = \mA \gT_g^x \vx + \varepsilon,
\end{equation}

where the sensing matrix $\mA$ is known and $g$ stands for the unknown rotation angle (element of the group of rotations). 
Our goal is to reconstruct the rotated signal $\gT_g^x \vx$. The way to address this problem depends on the type of prior generative model that we are using. Below we consider three different scenarios. 

\paragraph{Standard Prior}
Firstly, we may have a prior generative model, which is uninformed of the rotation. That is, it was trained on the signals in the non-rotated (canonical) orientation and it is only able to generate such. However, it is still possible to reformulate the optimization problem (\ref{eq:cs_mle}), so that we are able to reconstruct the rotated signal:

\begin{align}
    \vz^*, g^* &= \arg\min_{\vz, g} \|\vy - \mA \gT_{g}^x G(\vz)\|_2^2, \\
    \vx^* &= \gT_{g^*}^x G(\vz^*).
\end{align}

However, we have observed that optimizing for the latent code and the rotation angle simultaneously using the gradient descent does not produce reasonable results. However, coordinate gradient descent can be applied instead. In the latter case, we alternate the gradient steps to update the latent code and the rotation angle. This allows us to use different learning rates and considerably improves the convergence. 

\paragraph{Conditional Prior}
Secondly, we may have a conditional generative model as a prior. In this case, the model is trained on rotated images. As a result, it can generate a rotated image from a latent code and with the desired angle. The optimization task for an MLE solution will then be the following:
\begin{align}
    \vz^*, g^* &= \arg\min_{\vz, g} \|\vy - \mA G(\vz, g)\|_2^2, \\
    \vx^{*} &= G(\vz^*, g^*).
\end{align}

\paragraph{Prior with rotation-aware latent space}
Finally, we may train a prior with a latent code for all the rotated images. For example, we can augment the dataset with the rotated sample while training the prior. Another option would be to train an equivariant generative model. The latter approach does not require the rotated samples during training. However, the equivariance property still allows it to produce the rotated samples. 

\begin{align}
    \vz^* &= \arg\min_{\vz} \|\vy - \mA G(\vz)\|_2^2,\\
    \vx^{*} &= G(\vz^*).
\end{align}

The straightforward advantage of this approach is that we have the same objective as a conventional generative compressed sensing and there is no need to optimize with respect to the rotation angle. Furthermore, in case of equivariant prior, we do not need any additional (augmented) data while training the generative model.

\subsection{Recovery guarantees for equivariant priors}
With some conditions on the sensing matrix, generative  compressed sensing has recovery guarantees in classical setup. The main idea is presented in \cite{Bora2017-as} with follow-up works on more general sensing setups \cite{liu_generalized_2020, liu_sample_2020}. We can provide the same recovery guarantees for equivariant priors. Before stating the result, we introduce some notations. A generative model $G(\cdot)$ is defined as a mapping from $k$-dimensional ball of radius $r$ in $\R^k$ to $\R^n$. We assume that the action of rotation in $\R^n$ is given by a unitary representation ${\gT_g^x}$.

\begin{theorem}
Consider an equivariant generative model $G: B_k(r)\to\R^n$, which is assumed to be an $L$-Lipschitz function. Let $\mA\in\R^{m\times n}$ be a random Gaussian matrix with i.i.d. entries $\gN(0,m^{-1})$. Assume that a signal $\vx^\star\in\R^n$ is observed in an unknown orientation according to $\vy = \mA \gT_g^x \vx_\star + \vepsilon$. Suppose that $\hat{\vz}$ minimizes $\norm{\vy-\mA G(\vz)}$ to within error of $\delta_{\text{approx}}$.
For measurement  number $m =\Omega(k \log (Lr/\delta)$, the following holds with probability $1-e^{-\Omega(m)}$
\begin{align*}
&\norm{G(\hat{\vz})-\gT_g^x \vx_\star}_2  \leq  \\
& \quad\quad 6 \min_{\vz\in B_k(r)}\norm{G(\vz)-\vx_\star}_2 + 3\norm{\vepsilon}_2 + 2\delta_{\text{approx}} + 2\delta.
\end{align*}
\label{thm:equivmodel}
\end{theorem}
Since equivariant models incorporate the rotation in the latent space, the proof remains the same. Nonetheless, we review it in the Appendix.

Note that the upper bound on the reconstruction error does not depend on the rotation operation $\gT_g^x$. The term $\min_{\vz\in B_k(r)}\norm{G(\vz)-\vx_\star}_2$ is evaluated only on the canonical pose. This is a built-in feature of the equivariant models, namely that,:
\begin{align*}
\norm{G(\vz)-\gT_g^x \vx_\star}_2 &= \norm{{\gT_g^x}^{H}G(\vz)- \vx_\star}_2\\
&= \norm{G({\gT_g^z}^{H}\vz)- \vx_\star}_2.
\end{align*}
Since ${\gT_g^z}$ is a unitary matrix, it maps $B_k(r)$ to itself, and ${\gT_g^z}^{H}\vz$ remains inside $B_k(r)$. This means that:
\begin{align*}
\min_{\vz\in B_k(r)}\norm{G(\vz)-\gT_g^x \vx_\star}_2 
&=\min_{\vz\in B_k(r)} \norm{G({\gT_g^z}^{H}\vz)- \vx_\star}_2\\
&=\min_{\vz\in B_k(r)} \norm{G(\vz)- \vx_\star}_2.
\end{align*}
This upper bound remains the same for generative compressed sensing with no rotation. Although this is just an upper bound, it suggests that there should not be a difference in the performance of equivariant priors on rotated and non-rotated data. Our experimental results confirm that there is a small difference in performance for rotated and non-rotated signals. 
\section{Experiments}\label{sec:experiments}

We consider two different setups. The conventional compressed sensing discussed in section \ref{sec:related_gen_cs} (no rotation) and compressed sensing with unknown orientation discussed in section \ref{sec:gen_cs_rotation} (unknown rotation).  

\paragraph{Datasets}
We conduct experiments on two different datasets. We start with benchmarking experiments on MNIST. 
Subsequently, concerning a real-world application of the proposed approach, we conduct experiments on the Low Dose CT Image and Projection Data\footnote{\url{https://www.aapm.org/grandchallenge/lowdosect/}} (MAYO) dataset \citep{moen2021low}, which consist of three types of data: DICOM-CT-PD projection data, DICOM image data, and Excel clinical data reports. To our aim, we use the DICOM subset only. Images are divided into three sets labelled N for neuro, C for chest, and L for liver each of which comprises 512x512 images from 50 different patients. To train the generative priors, we consider the L subset which is made of $\sim$7K samples that we divide into train, validation and test sets comprising $\sim$80\%, $\sim$10\%, and $\sim$10\% of the images, respectively.
Before feeding a model, we randomly crop the image and then rescale it to 128x128, and finally, we normalize the pixels value in $[0,1]$.

\begin{figure}[t]
    \centering
    \begin{tabular}{l}
        \multicolumn{1}{c}{\Large{No rotation}}\\
        \includegraphics[width=0.39\textwidth]{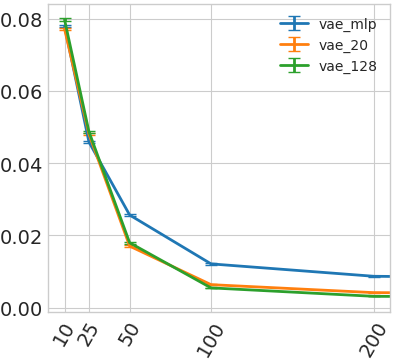} \\
        \multicolumn{1}{c}{Number of measurements}\\
        \text{}\\
         \multicolumn{1}{c}{\Large{Unknown rotation}}\\
        \includegraphics[width=0.39\textwidth]{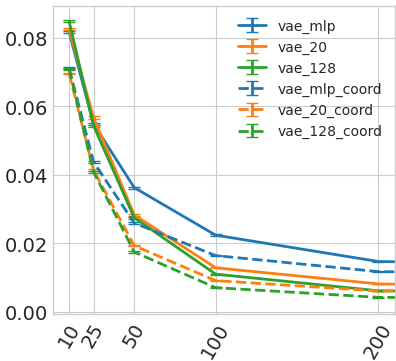} \\
        \multicolumn{1}{c}{Number of measurements}\\
    \end{tabular}
    \caption{Comparison of MLP-VAE and Convolutional VAE with the latent dimentions 20 and 128 on MNIST dataset. Average MSE for 10000 test points.}
    \label{fig:vae_prior_types}
    \vskip -0.1in
\end{figure}

\paragraph{Prior Generative Models}
We compare various generative models as prior on the signals that we aim to reconstruct. Broadly speaking, we consider two types of generative models in our experiments: VAE and Normalizing Flow. In the former case, we have MLP-VAE, fully convolutional VAE (Conv-VAE) and Equivariant VAE (Eq-VAE: proposed model). And in the latter, as a flow prior, we consider a multi-scale RealNVP model. To bring the non-equivariant models, i.e., flow and VAE, at par with the Eq-VAE in terms of their ability to generate rotated images so that they can be effectively used as a prior for CS with unknown rotation, we consider two routes, 1. Conditional model: we include a conditional flow (Cond. Flow) model, which has rotation angle as an additional input and is trained on the augmented (rotated) dataset, 2. Augmented models: another way to augment the capabilities of conventional generative models is to train them on augmented (rotated) datasets. On this line, we include augmented VAE (Aug-VAE), and augmented flow (Aug-Flow) as conventional VAE and Flow but explicitly trained on augmented (addition of random rotation as a pre-processing step) dataset. In our MNIST experiments, we noticed impractically high latency of Cond. Flow during both the generative prior training and the following CS experiment. The problem gets highly exacerbated for the MAYO dataset as it is more complex than MNIST, and hence, we decided to drop the Cond. Flow for MAYO due to its impractical latency.


\paragraph{Metrics}
When it comes to measuring the success of the compressed sensing, we are interested in two criteria: reconstruction quality and convergence speed. We measure MSE to evaluate the quality of the reconstruction. For convergence, we assume that if the MSE (per pixel) is lower than 0.01, then the reconstruction is successful. We then report the proportion of converged points and the average number of iterations (or gradient descent steps) required for a method to converge. 

\begin{table*}[ht]
\caption{CS results on MNIST. We report results averaged over 1000 test points. By Converged Points, we refer to points that, for a given number of iterations, resulted in MSE (per pixel) lower than $1.e^{-2}$. We emphasize in bold, results from the best model.}

\label{tab:mnist_cs_results}
\begin{center}
\begin{tabularx}{\linewidth}{@{}lll|YYc|YYc|YYc@{}}
\toprule
&&\multirow{2}{*}{Prior} & \multicolumn{3}{c|}{50 measurements} & \multicolumn{3}{c|}{100 measurements} & \multicolumn{3}{c}{200 measurements} \\
&&& MSE ($1.e^{-3}$)$\downarrow$ & Converged Points (\%)$\uparrow$ & \# iter$\downarrow$ &  MSE ($1.e^{-3}$)$\downarrow$ & Converged Points (\%)$\uparrow$ & \# iter$\downarrow$ &  MSE ($1.e^{-3}$)$\downarrow$ & Converged Points (\%)$\uparrow$ & \# iter $\downarrow$\\ \midrule
&\multirow{5}{*}{\STAB{\rotatebox[origin=c]{90}{No rotation}}}
& Flow        & 45.4 & 100 & 59 & 19.2 & 100 & 39 & 10.2 & 100 & 38 \\
&& Cond. Flow & 22.6 & 100 & 35 & 10.1 & 100 & \textbf{22} & 5.8 & 100 & \textbf{22} \\
&& MLP-VAE    & 25.7 & 100 & 50 & 12.1 & 100 & 41 & 8.7 & 100 & 38\\
&& Conv-VAE   & 18.0 & 100 & 48 & 5.5  & 100 & 39 & 3.2 & 100 & 35\\
&& Eq-VAE     & \textbf{10.6} & 100 & \textbf{34} & \textbf{2.6}  & 100 & \textbf{22} & \textbf{2.1} & 100 & \textbf{22}\\ \midrule
\multirow{5}{*}{\STAB{\rotatebox[origin=c]{90}{Unknown}}}
& \multirow {5}{*}{\STAB{\rotatebox[origin=c]{90}{rotation}}}
& Flow        & 27.2 & 100 & 73 & 13.2 & 100 & 60 & 7.6  & 100 & 36 \\
&& Cond. Flow & 17.9 & 100 & 42 & 7.2  & 100 & 32 & 6.0  & 100 & 30 \\
&& MLP-VAE    & 25.9 & 100 & 48 & 16.4 & 100 & 40 & 11.7 & 100 & 34\\
&& Conv-VAE   & 17.5 & 100 & 51 & 7.0  & 100 & 39 & 4.2  & 100 & 32\\
&& Eq-VAE     & \textbf{11.8} & 100 & \textbf{33} & \textbf{4.2}  & 100 & \textbf{24} & \textbf{3.2}  & 100 & \textbf{23}\\
\bottomrule
\end{tabularx}
\end{center}
\end{table*}

\subsection{Benchmark experiments on MNIST}
Concerning compressed sensing experiments, we consider a different number of linear measurements to reconstruct the input signal. Specifically, in \autoref{tab:mnist_cs_results} we report CS results for the MNIST dataset considering 50, 100, and 200 measurements. For each of the mentioned values, and each prior model, we report the average value for the MSE and the number of iterations required to achieve a 100\% success rate for reconstructing the input signals.
As mentioned previously, we consider two different scenarios: no rotations and unknown rotations. In both cases, Eq-VAE prior report the best performance concerning MSE and the number of iterations required to reconstruct the input images.
Indeed, although with all the priors we observe a 100\% convergence rate for the CS experiments, Eq-VAE requires fewer iterations and reports a lower average MSE than all the other models.

Following \cite{Bora2017-as} we train MLP-VAE with 20 dimensional latent space. However, as can be seen from the Figure \ref{fig:vae_prior_types}, fully convolutional VAE with the same latent space dimensionality tends to perform better both with and without rotation. Presumably, the introduction of the fully convolutional architecture makes the model equivariant to the translation, which is beneficial for the compressed sensing. As results in Table \ref{tab:mnist_cs_results} show, equivariance to rotation improves the performance even further. 

Figure \ref{fig:vae_prior_types} also shows that larger latent space of the convolutional VAE gives a slight performance improvement. Thus we chose to train equivariant prior with the same latent dimension. When rotation is unknown, we also compare coordinate gradient descent (\textit{dashed line}) with the joint optimization of the latent code and the rotation angle (\textit{solid line}). In all the experiments, coordinate gradient descent shows better results.

\begin{table*}[!ht]
\caption{CS results on MAYO. The reported results corresponding to 200 measurements. By Converged Points, we refer to points that, for a given number of iterations, resulted in MSE (per pixel) lower than $1.e^{-2}$. Concerning the MSE, we reported the mean value $\pm 1 \sigma$. The MSE is computed across converged points only. We emphasize in bold, results from the best model.}

\label{tab:mayo_cs_results}
\begin{center}
\begin{tabularx}{\linewidth}{@{}lll|YY|YY|YY|YY@{}}
\toprule
&&\multirow{2}{*}{Prior} & \multicolumn{2}{c|}{50 iterations} & \multicolumn{2}{c|}{100 iterations} & \multicolumn{2}{c|}{150 iterations} & \multicolumn{2}{c}{200 iterations} \\
&&& MSE ($1.e^{-3}$)$\downarrow$ & Converged Points (\%)$\uparrow$  &  MSE ($1.e^{-3}$)$\downarrow$ & Converged Points (\%)$\uparrow$  &  MSE ($1.e^{-3}$)$\downarrow$ & Converged Points (\%)$\uparrow$  &  MSE ($1.e^{-3}$)$\downarrow$ & Converged Points (\%)$\uparrow$ \\ \midrule

&\multirow{5}{*}{\STAB{\rotatebox[origin=c]{90}{No rotation}}}
& Flow & 8.9 $\pm$ 0.9 & 28.2 & 8.1 $\pm$ 1.2 & 65.9 & 7.3 $\pm$ 1.3 & 79.7 & 6.8 $\pm$ 1.4  & 86.8  \\
&& Aug-Flow & N/A & 0.0 & 8.2 $\pm$ 1.2 & 40.1 & 7.4 $\pm$ 1.4 & 77.1 & 6.7 $\pm$ 1.4 & 83.4 \\
&& Conv-VAE & 7.9 $\pm$ 1.3 & 39.9 & 6.8 $\pm$ 1.6 & 76.2 & 6.2 $\pm$ 1.8 & \bf{89.6} & 5.9 $\pm$ 1.8  & \bf{95.5}  \\
&& Aug-VAE & 7.5 $\pm$ 1.3 & 51.7 & \bf{6.1 $\pm$ 1.6} & 75.7 & \bf{5.7 $\pm$ 1.7} & 83.8 & \bf{5.5 $\pm$ 1.6}  & 88.1  \\
&& Eq-VAE &  \bf{7.4 $\pm$ 1.4} & \bf{73.5} & 6.4 $\pm$ 1.4 & \bf{86.9} & 6.1 $\pm$ 1.5  & 89.2  & 6.0 $\pm$ 1.5  & 90.4  \\ \midrule
\multirow{5}{*}{\STAB{\rotatebox[origin=c]{90}{Unknown}}}
& \multirow{5}{*}{\STAB{\rotatebox[origin=c]{90}{rotation}}}
& Flow & 7.2 $\pm$ 1.4 & \bf{76.4} & \bf{5.8 $\pm$ 1.5} & 83.8 & \bf{5.2 $\pm$ 1.6} & 87.5 & \bf{5.0 $\pm$ 1.7} & \bf{93.3}  \\
&& Aug-Flow & N/A & 0.0 & 8.2 $\pm$ 1.1 & 49.9 & 6.9 $\pm$ 1.3 & 80.9 & 6.0 $\pm$ 1.3 & 84.3 \\
&& Conv-VAE & \bf{7.1 $\pm$ 1.7} & 55.3 & 5.9 $\pm$ 2.1 & 80.5 & 5.4 $\pm$ 2.1 & 88.2 & 5.0 $\pm$ 2.0  & 90.2 \\
&& Aug-VAE & 7.5 $\pm$ 1.5 & 59.3 & 5.9 $\pm$ 1.6 & 85.2 & 5.3 $\pm$ 1.5 & 89.1 & 5.1 $\pm$ 1.6  & 92.4  \\
&& Eq-VAE &  7.6 $\pm$ 1.4 & 70.3 & 6.6 $\pm$ 1.5 & \bf{87.5} & 6.1 $\pm$ 1.5 & \bf{91.3} & 6.0 $\pm$ 1.6 & 92.7  \\ \midrule
\bottomrule
\end{tabularx}
\end{center}
\end{table*}

\subsection{CT scans reconstruction: MAYO dataset}
We report in~\autoref{tab:mayo_cs_results} results for the MAYO dataset. Differently from the MNIST result, in this case, we fix the number of measurements to 200 and evaluate the performance of the different generative priors on the CS task considering a different number of iterations. Specifically, we consider 50, 100, 150, and 200 iterations, and for each of them, we report the average MSE and the percentage of points successfully reconstructed (i.e., with MSE per pixel below 0.01). 
Being the MAYO dataset more complex than the MNIST, in this case, we do not observe a clear dominance of one type of prior compared to the others. However, as a general remark, we can notice that using Eq-VAE the average percentage of converged points is higher than for other models for both no rotations, 85.0\%, and unknown rotations, 85.5\%. As a comparison, the second-best performing priors are the VAE with 75.3\%, and the Flow with 85.2\%, concerning no and unknown rotations, respectively. 
Instead, regarding the MSE, all the results agree within 1 standard deviation.

\begin{figure*}[!ht]
    \centering
    \includegraphics[width=\linewidth]{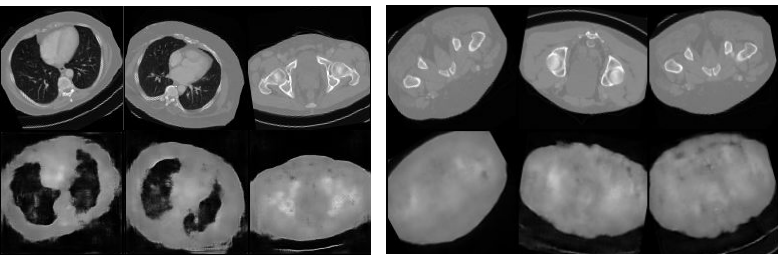}\hfill
    \caption{Example of reconstructed images using Eq-VAE (left-3) and VAE (right-3) as a prior. The top row shows ground truth rotated images prior to measurement, and the bottom row shows the corresponding CS reconstructed images. We notice in the third column (both left and right images), the VAE reconstructed image is $180^{\circ}$ rotated with respect to the canonical configuration while the Eq. VAE is able to retrieve the configuration effectively.}
    \label{fig:mayo_eqvae_vae_rec_unk}
\end{figure*}

\subsection{Discussion}
As we discussed before, the upper bound on the reconstruction error in Theorem \ref{thm:equivmodel} suggests that there is no difference in the performance of equivariant models in rotated and non-rotated cases. The results for the MAYO dataset show a small performance drop from no rotation to unknown rotation cases. The discrepancy is a bit larger for MNIST experiments. 
The difference can be attributed to the fact that equivariant models are built for finite rotation groups while we test on continuous rotation groups. 
In Figure~\ref{fig:mayo_eqvae_vae_rec_unk} we report a comparison between reconstructed rotated samples from Eq-VAE(left) and VAE (right). As expected, Eq-VAE successfully recovers the correct orientation for the reconstructed images while VAE struggles to achieve such a goal. 

\paragraph{Group Choice} 
We train the equivariant prior under the action of the cyclic group. This is a discrete group, size of which, is a hyperparameter to choose. However, in our compressed sensing experiments with unknown orientation, the rotation angle is sampled uniformly. Thus, a larger group size will potentially lead to a better-compressed sensing performance. We compared three different group sizes (see Appendix \ref{appx:eq_vae_group}) and conclude that group $C_{16}$ results in the best reconstruction results in both the CS setups.

\section{Conclusion}
We consider a generative compressed sensing problem with unknown orientation of the measurement signal. This new setup motivates the usage of new generative prior models, which are capable of producing rotated images. We propose an equivariant variational autoencoder and extend theoretical convergence guarantees to the case of unknown rotation and equivariant generative prior. We experimentally show that the proposed equivariant prior is better or comparable to other benchmarks in terms of reconstruction quality and provide potential additional advantages in terms of convergence speed, as it usually requires fewer iterations (gradient descent steps).


\bibliography{main}

\begin{thebibliography}{36}
\providecommand{\natexlab}[1]{#1}
\providecommand{\url}[1]{\texttt{#1}}
\expandafter\ifx\csname urlstyle\endcsname\relax
  \providecommand{\doi}[1]{doi: #1}\else
  \providecommand{\doi}{doi: \begingroup \urlstyle{rm}\Url}\fi

\bibitem[Asim et~al.(2019)Asim, Daniels, Leong, Ahmed, and Hand]{Asim2019-lv}
Asim, M., Daniels, M., Leong, O., Ahmed, A., and Hand, P.
\newblock Invertible generative models for inverse problems: mitigating
  representation error and dataset bias.
\newblock \emph{arXiv:1905.11672}, May 2019.

\bibitem[Bendory et~al.(2017)Bendory, Boumal, Ma, Zhao, and
  Singer]{bendory2017bispectrum}
Bendory, T., Boumal, N., Ma, C., Zhao, Z., and Singer, A.
\newblock Bispectrum inversion with application to multireference alignment.
\newblock \emph{IEEE Transactions on signal processing}, 66\penalty0
  (4):\penalty0 1037--1050, 2017.

\bibitem[Bepler et~al.(2019)Bepler, Zhong, Kelley, Brignole, and
  Berger]{Bepler2019-qi}
Bepler, T., Zhong, E.~D., Kelley, K., Brignole, E., and Berger, B.
\newblock Explicitly disentangling image content from translation and rotation
  with {spatial-VAE}.
\newblock \emph{arXiv:1909.11663}, September 2019.

\bibitem[Bora et~al.(2017)Bora, Jalal, Price, and Dimakis]{Bora2017-as}
Bora, A., Jalal, A., Price, E., and Dimakis, A.~G.
\newblock Compressed sensing using generative models.
\newblock In Precup, D. and Teh, Y.~W. (eds.), \emph{Proceedings of the 34th
  International Conference on Machine Learning}, volume~70 of \emph{Proceedings
  of Machine Learning Research}, pp.\  537--546. PMLR, 2017.

\bibitem[Candès et~al.(2006)Candès, Romberg, and Tao]{candes_stable_2006}
Candès, E.~J., Romberg, J.~K., and Tao, T.
\newblock Stable signal recovery from incomplete and inaccurate measurements.
\newblock \emph{Communications on Pure and Applied Mathematics}, 59\penalty0
  (8):\penalty0 1207--1223, 2006.

\bibitem[Cesa et~al.(2022)Cesa, Lang, and Weiler]{cesa2022a}
Cesa, G., Lang, L., and Weiler, M.
\newblock A program to build e(n)-equivariant steerable {CNN}s.
\newblock In \emph{International Conference on Learning Representations}, 2022.

\bibitem[Chen et~al.(2008)Chen, Tang, and Leng]{chen2008prior}
Chen, G.-H., Tang, J., and Leng, S.
\newblock Prior image constrained compressed sensing (piccs): a method to
  accurately reconstruct dynamic ct images from highly undersampled projection
  data sets.
\newblock \emph{Medical physics}, 35\penalty0 (2):\penalty0 660--663, 2008.

\bibitem[Cohen \& Welling(2016{\natexlab{a}})Cohen and Welling]{cohen2016group}
Cohen, T. and Welling, M.
\newblock Group equivariant convolutional networks.
\newblock In \emph{International conference on machine learning}, pp.\
  2990--2999. PMLR, 2016{\natexlab{a}}.

\bibitem[Cohen \& Welling(2016{\natexlab{b}})Cohen and
  Welling]{cohen_steerable_2016}
Cohen, T.~S. and Welling, M.
\newblock Steerable {CNNs}.
\newblock In \emph{{ICLR} 2017}, November 2016{\natexlab{b}}.

\bibitem[Daras et~al.(2021)Daras, Dean, Jalal, and
  Dimakis]{daras_intermediate_2021}
Daras, G., Dean, J., Jalal, A., and Dimakis, A.~G.
\newblock Intermediate {Layer} {Optimization} for {Inverse} {Problems} using
  {Deep} {Generative} {Models}.
\newblock \emph{arXiv: 2102.07364}, February 2021.
\newblock URL \url{http://arxiv.org/abs/2102.07364}.

\bibitem[Davenport et~al.(2007)Davenport, Duarte, Wakin, Laska, Takhar, Kelly,
  and Baraniuk]{davenport2007smashed}
Davenport, M.~A., Duarte, M.~F., Wakin, M.~B., Laska, J.~N., Takhar, D., Kelly,
  K.~F., and Baraniuk, R.~G.
\newblock The smashed filter for compressive classification and target
  recognition.
\newblock In \emph{Computational Imaging V}, volume 6498, pp.\  64980H.
  International Society for Optics and Photonics, 2007.

\bibitem[Donoho(2006)]{donoho_compressed_2006}
Donoho, D.~L.
\newblock Compressed sensing.
\newblock \emph{IEEE Transactions on Information Theory}, 52\penalty0
  (4):\penalty0 1289--1306, April 2006.

\bibitem[Duarte et~al.(2008)Duarte, Davenport, Takhar, Laska, Sun, Kelly, and
  Baraniuk]{duarte2008single}
Duarte, M.~F., Davenport, M.~A., Takhar, D., Laska, J.~N., Sun, T., Kelly,
  K.~F., and Baraniuk, R.~G.
\newblock Single-pixel imaging via compressive sampling.
\newblock \emph{IEEE signal processing magazine}, 25\penalty0 (2):\penalty0
  83--91, 2008.

\bibitem[Falorsi et~al.(2018)Falorsi, de~Haan, Davidson, De~Cao, Weiler,
  Forr{\'e}, and Cohen]{falorsi2018explorations}
Falorsi, L., de~Haan, P., Davidson, T.~R., De~Cao, N., Weiler, M., Forr{\'e},
  P., and Cohen, T.~S.
\newblock Explorations in homeomorphic variational auto-encoding.
\newblock \emph{ICML workshop on Theoretical Foundations and Applications of
  Deep Generative Models}, 2018.

\bibitem[Foucart \& Rauhut(2013)Foucart and Rauhut]{foucart_mathematical_2013}
Foucart, S. and Rauhut, H.
\newblock \emph{A {Mathematical} {Introduction} to {Compressive} {Sensing}}.
\newblock Applied and {Numerical} {Harmonic} {Analysis}. Springer New York, New
  York, NY, 2013.

\bibitem[Hussein et~al.(2020)Hussein, Tirer, and Giryes]{Hussein2020-yk}
Hussein, S.~A., Tirer, T., and Giryes, R.
\newblock {Image-Adaptive} {GAN} based reconstruction.
\newblock \emph{AAAI}, 34\penalty0 (04):\penalty0 3121--3129, April 2020.

\bibitem[Jordan et~al.(1999)Jordan, Ghahramani, Jaakkola, and
  Saul]{jordan1999introduction}
Jordan, M.~I., Ghahramani, Z., Jaakkola, T.~S., and Saul, L.~K.
\newblock An introduction to variational methods for graphical models.
\newblock \emph{Machine learning}, 37\penalty0 (2):\penalty0 183--233, 1999.

\bibitem[Karras et~al.(2021)Karras, Aittala, Laine, H{\"a}rk{\"o}nen, Hellsten,
  Lehtinen, and Aila]{Karras2021-yx}
Karras, T., Aittala, M., Laine, S., H{\"a}rk{\"o}nen, E., Hellsten, J.,
  Lehtinen, J., and Aila, T.
\newblock {Alias-Free} generative adversarial networks.
\newblock \emph{arXiv:2106.12423}, June 2021.

\bibitem[Kingma \& Welling(2013)Kingma and Welling]{kingma2013auto}
Kingma, D.~P. and Welling, M.
\newblock Auto-encoding variational bayes.
\newblock \emph{arXiv preprint arXiv:1312.6114}, 2013.

\bibitem[K{\"o}hler et~al.(2020)K{\"o}hler, Klein, and
  No{\'e}]{kohler2020equivariant}
K{\"o}hler, J., Klein, L., and No{\'e}, F.
\newblock Equivariant flows: exact likelihood generative learning for symmetric
  densities.
\newblock In \emph{International Conference on Machine Learning}, pp.\
  5361--5370. PMLR, 2020.

\bibitem[Kondor \& Trivedi(2018)Kondor and Trivedi]{Kondor2018-GENERAL}
Kondor, R. and Trivedi, S.
\newblock On the generalization of equivariance and convolution in neural
  networks to the action of compact groups.
\newblock In \emph{International Conference on Machine Learning (ICML)}, 2018.

\bibitem[Lei et~al.(2019)Lei, Jalal, Dhillon, and Dimakis]{lei_inverting_2019}
Lei, Q., Jalal, A., Dhillon, I.~S., and Dimakis, A.~G.
\newblock Inverting {Deep} {Generative} models, {One} layer at a time.
\newblock In \emph{Advances in {Neural} {Information} {Processing} {Systems}},
  pp.\  13910--13919, 2019.

\bibitem[Liu \& Scarlett(2020)Liu and Scarlett]{liu_generalized_2020}
Liu, Z. and Scarlett, J.
\newblock The {Generalized} {Lasso} with {Nonlinear} {Observations} and
  {Generative} {Priors}.
\newblock In Larochelle, H., Ranzato, M., Hadsell, R., Balcan, M.~F., and Lin,
  H. (eds.), \emph{Advances in {Neural} {Information} {Processing} {Systems}},
  volume~33, pp.\  19125--19136. Curran Associates, Inc., 2020.

\bibitem[Liu et~al.(2020)Liu, Gomes, Tiwari, and Scarlett]{liu_sample_2020}
Liu, Z., Gomes, S., Tiwari, A., and Scarlett, J.
\newblock Sample {Complexity} {Bounds} for 1-bit {Compressive} {Sensing} and
  {Binary} {Stable} {Embeddings} with {Generative} {Priors}.
\newblock In III, H.~D. and Singh, A. (eds.), \emph{Proceedings of the 37th
  {International} {Conference} on {Machine} {Learning}}, volume 119 of
  \emph{Proceedings of {Machine} {Learning} {Research}}, pp.\  6216--6225.
  PMLR, July 2020.

\bibitem[Lustig et~al.(2007)Lustig, Donoho, and Pauly]{lustig2007sparse}
Lustig, M., Donoho, D., and Pauly, J.~M.
\newblock Sparse mri: The application of compressed sensing for rapid mr
  imaging.
\newblock \emph{Magnetic Resonance in Medicine: An Official Journal of the
  International Society for Magnetic Resonance in Medicine}, 58\penalty0
  (6):\penalty0 1182--1195, 2007.

\bibitem[Moen et~al.(2021)Moen, Chen, Holmes~III, Duan, Yu, Yu, Leng, Fletcher,
  and McCollough]{moen2021low}
Moen, T.~R., Chen, B., Holmes~III, D.~R., Duan, X., Yu, Z., Yu, L., Leng, S.,
  Fletcher, J.~G., and McCollough, C.~H.
\newblock Low-dose ct image and projection dataset.
\newblock \emph{Medical physics}, 48\penalty0 (2):\penalty0 902--911, 2021.

\bibitem[Paredes et~al.(2007)Paredes, Arce, and Wang]{paredes2007ultra}
Paredes, J.~L., Arce, G.~R., and Wang, Z.
\newblock Ultra-wideband compressed sensing: Channel estimation.
\newblock \emph{IEEE Journal of Selected Topics in Signal Processing},
  1\penalty0 (3):\penalty0 383--395, 2007.

\bibitem[Rezende et~al.(2014)Rezende, Mohamed, and
  Wierstra]{rezende2014stochastic}
Rezende, D.~J., Mohamed, S., and Wierstra, D.
\newblock Stochastic backpropagation and approximate inference in deep
  generative models.
\newblock In \emph{International Conference on Machine Learning}, pp.\
  1278--1286. PMLR, 2014.

\bibitem[Satorras et~al.(2021)Satorras, Hoogeboom, Fuchs, Posner, and
  Welling]{satorras2021n}
Satorras, V.~G., Hoogeboom, E., Fuchs, F.~B., Posner, I., and Welling, M.
\newblock E (n) equivariant normalizing flows.
\newblock In \emph{Advances in Neural Information Processing Systems}, 2021.

\bibitem[Serre(1977)]{serre1977linear}
Serre, J.-P.
\newblock \emph{Linear representations of finite groups}.
\newblock Springer, 1977.

\bibitem[Singer(2018)]{singer_mathematics_2018}
Singer, A.
\newblock Mathematics for cryo-electron microscopy.
\newblock In \emph{Proceedings of the {International} {Congress} of
  {Mathematicians}: {Rio} de {Janeiro} 2018}, pp.\  3995--4014. World
  Scientific, 2018.

\bibitem[Tibshirani(1996)]{tibshirani1996regression}
Tibshirani, R.
\newblock Regression shrinkage and selection via the lasso.
\newblock \emph{Journal of the Royal Statistical Society: Series B
  (Methodological)}, 58\penalty0 (1):\penalty0 267--288, 1996.

\bibitem[Weiler \& Cesa(2019)Weiler and Cesa]{e2cnn}
Weiler, M. and Cesa, G.
\newblock {General E(2)-Equivariant Steerable CNNs}.
\newblock In \emph{Conference on Neural Information Processing Systems
  (NeurIPS)}, 2019.

\bibitem[Whang et~al.(2021{\natexlab{a}})Whang, Lei, and Dimakis]{Whang2021-fj}
Whang, J., Lei, Q., and Dimakis, A.
\newblock Solving inverse problems with a flow-based noise model.
\newblock In Meila, M. and Zhang, T. (eds.), \emph{Proceedings of the 38th
  International Conference on Machine Learning}, volume 139 of
  \emph{Proceedings of Machine Learning Research}, pp.\  11146--11157. PMLR,
  2021{\natexlab{a}}.

\bibitem[Whang et~al.(2021{\natexlab{b}})Whang, Lindgren, and
  Dimakis]{Whang2021-if}
Whang, J., Lindgren, E., and Dimakis, A.
\newblock Composing normalizing flows for inverse problems.
\newblock In Meila, M. and Zhang, T. (eds.), \emph{Proceedings of the 38th
  International Conference on Machine Learning}, volume 139 of
  \emph{Proceedings of Machine Learning Research}, pp.\  11158--11169. PMLR,
  2021{\natexlab{b}}.

\bibitem[Whang et~al.(2021{\natexlab{c}})Whang, Lindgren, and
  Dimakis]{Whang2021-ro}
Whang, J., Lindgren, E., and Dimakis, A.
\newblock Composing normalizing flows for inverse problems.
\newblock In Meila, M. and Zhang, T. (eds.), \emph{Proceedings of the 38th
  International Conference on Machine Learning}, volume 139 of
  \emph{Proceedings of Machine Learning Research}, pp.\  11158--11169. PMLR,
  2021{\natexlab{c}}.

\end{thebibliography}
\bibliographystyle{icml2022}

\newpage
\onecolumn
\appendix
\section{Group Theory and Equivariant Neural Networks}

We review some basics of group theory and group convolutional neural networks. There are other ways of building equivariant networks beyond group convolutional neural networks, see for example \cite{e2cnn,cohen_steerable_2016,Kondor2018-GENERAL,cesa2022a}. Our theoretical results and equivariant VAE constructions are oblivious to the choice of architecture. 

For a set $\gG$, a law of composition $\cdot : \gG \times \gG \to \gG$ is a mapping that maps $h, g\in\gG$ to $h \cdot g\in\gG$. A \textit{group} is a set $\gG$ with a  law of composition $\cdot : \gG \times \gG \to \gG$, called {group law},  which satisfies following properties:
    \begin{itemize}
        \item The law $\cdot$ is associative: $a \cdot (b \cdot c) = (a \cdot b) \cdot c$ for all $ a, b, c \in \gG$
        \item There is an identity element in $\gG$, denoted by $1$, such that $a \cdot 1 = 1 \cdot a = a$ for all $a\in\gG$
        \item Every element of $a\in\gG$ has an inverse, denoted by $b$ such that $a\cdot b=1$ and $b\cdot a=1$.
    \end{itemize}

A group $\gG$ is an abelian group if the group law is {commutative}, namely $a\cdot b =b \cdot a$. 
A group $\gG$ is {finite} if the set $\gG$ is finite. A group $\gG$ is a {compact} group if $\gG$ is a compact topological space with continuous group operation.

For an arbitrary space $\gX$, we can define the action of the group $\gG$ by a mapping $\gT_g: \gG \times \gX \to \gX$, which maps $\vx\in\gX$ and $g\in\gG$ to $\gT_g\vx$. For the identity element, $\vx$ is mapped to itself. In this work, the action of the group is assumed to be linear on a vector space $\gX$, and $\gT_g$ has a matrix representation. The action of the group $\gG$ satisfies group properties indicates above, and therefore, $\gT_g$ is a linear representation of the group $\gG$ on $\gX$. Representation theory characterizes linear representations of groups on general vector spaces. We work with unitary representations, that is,  $\gT_g$ is a unitary transformation.

A mapping $f:\gX\to\gY$ is equivariant to the action of the group $\gG$ if
\[
f(\gT^{\gX}_g\vx) =\gT^{\gY}_g f(\vx),
\]
where $\gT^{\gX}_g$ and $\gT^{\gY}_g$ are linear representations of the group $\gG$ on $\gX$ and $\gY$.

\paragraph{Group Convolution} One example of equivariant neural networks is  group equivariant convolutional networks \cite{cohen2016group}. 
The group convolution is defined on the space $\gX$ as:
\begin{align}
    \forall g \in \gG \quad  (\vw \gconv \vx)(g) := \langle \gT_g\vw, \vx \rangle, 
\label{def:group_conv}
\end{align}
where $\vw,\vx\in\gX$. The group convolution is an equivariant transformation w.r.t. the actions of group $\gG$. Note that $(\vw \gconv \vx)\in\R^{|\gG|}$, and the actions of group $\gG$ on $\vx\in\R^{|\gG|}$ is permuting the entries of $\vx$. This means that the regular representation $\gG$ is used as group representation \cite{serre1977linear}. 

Each group convolution can be seen as a function defined on the group $\gG$, and therefore, after the first layer, we can focus on functions on $\gG$ determined by $|\gG|$-dimensional vectors. 
Given the group $\gG=(g_0, g_1, \dots, g_i, \dots, g_{|G|-1})$, and vectors $\vx, \vw \in \R^{|\gG|}$, the group convolution can be defined by a matrix multiplication between the vector $\vx$ and a matrix $\mW$:
\begin{equation}
    (\vw \gconv \vx)(g_i) = (\mW\vx)[i]
\end{equation}
where the matrix $\mW$ is given as follows:
\begin{equation}
    \mW=\begin{pmatrix}
    w(g_0)& \dots & w(g_{|\gG|-1})\\
    w(g_1^{-1}g_0) & \dots&   w(g_1^{-1}g_{|\gG|-1})\\
    \vdots &\ddots &\vdots \\
    w(g_{|\gG|-1}^{-1}g_0) & \dots&   w(g_{|\gG|-1}^{-1}g_{|\gG|-1}).\\
    \end{pmatrix}
\label{eq:group_circulant}
\end{equation}
A group convolutional neural network is defined by concatenating group convolution layers with pointwise non-linearities in the middle. At each layer, multiple 
kernel like $\vw$ can be chosen, each one corresponding to a single channel. It is possible to use other representations of the group $\gG$ to build equivariant networks (see \cite{cohen_steerable_2016,e2cnn} for some examples). 

\section{Recovery Guarantees and Theoretical Analysis} \label{appx:theory}
In this section, we present more details on theoretical guarantees for unknown rotation. We consider the following inverse problem with unknown rotation $g$: 
\begin{equation}
    \vy = \mA \gT_g^x \vx_\star + \vepsilon,
\end{equation}
where $\gT_g^x$ is the group representation in the input space. We work with unitary representations, which means that $(\gT_g^x)^{-1}=(\gT_g^x)^{H}$. We start by introducing some preliminary definitions.

\subsection{Definitions}
As we will see, we require the sensing matrix $\mA$ to satisfy certain constraints. The first on is set-restricted eigenvalue (S-REC) condition.
\begin{definition} A matrix $\mA$ is said to satisfy 
$\srec(\gS,\gamma,\delta)$ for some $\delta\geq 0$ and $\gamma>0$ and a set $\gS$ if for all $\vx,\vy\in\gS$:
\[
\norm{\mA(\vx-\vy)}_2\geq \gamma\norm{\vx-\vy}-\delta.
\]
\end{definition}
This is reminiscent of the null space property in compressed sensing. The idea is to guarantee that two points from the set of interest $S$ are \textit{separated} enough after mapping $\mA$, i.e., their difference lies away from the null space of $\mA$. 

The second condition on $\mA$ is that for every fixed $\vx\in\R^n$, we have:
\begin{equation}
    \norm{\mA\vx}_2\leq 2\norm{\vx}_2.
    \label{eq:norm_condition}
\end{equation}
Note that this bound needs to hold for a fixed $\vx$ and not all $\vx$, and therefore, this is a weaker assumption that bounding the spectral norm. We will see how these conditions are satisfied for a random Gaussian matrix.




\subsection{Proof}
To train the generative prior, we assume that the data is given in a canonical orientation. The existence of a canonical orientation is not necessary. The following argument can be reworked by considering the quotient space and its proper parametrization instead. Without rotation, generative priors enjoy recovery guarantees of type provided in \cite{Bora2017-as}. For canonical orientations, the very same analysis holds for equivariant generative priors too.  The difference appears for the scenario with unknown orientation $g$. 

The proof for the equivariant priors follows standard steps exactly as \cite{Bora2017-as}. The reason is that unknown rotations are absorbed into the latent of the generative model so the main steps remain intact. We replicate the essence of the proof here to be self-contained.

The proof strategy for Theorem \ref{thm:equivmodel} is as follows. First, let $\hat{\vz}$ be the estimated latent that minimizes $\norm{\vy-\mA G(\vz)}$ to within error of $\delta_{\text{approx}}$. This means that for all $\vz$, we have
\[
\norm{\vy-\mA G(\hat{\vz})}_2\leq \norm{\vy-\mA G({\vz})}_2 +\delta_{\text{approx}}.
\]
Also, let $\vz_\star$ be the latent code\footnote{Existence of $\hat{\vz}$ and $\vz_\star$ follows from compactness of domain and continuity of the objective function. For a more general case, the proof can be replicated by replacing $\hat{\vz}$ and $\vz_\star$ with their definition.} minimizing $\norm{\vx_\star- G(\vz)}_2$. Note that the equivariance condition implies that:
\[
\gT_g^x G(\vz) =  G(\gT_g^z\vz),
\]
for group representation $\gT_g^x$ and $\gT_g^z$. Therefore, the unitary property of $\gT_g^x$ implies that 
\[
\norm{\vx_\star- G(\vz)}_2 = \norm{\gT_g^x\vx_\star- \gT_g^x G(\vz)}_2 = 
 \norm{\gT_g^x\vx_\star- G(\gT_g^z\vz)}_2.
\]
This means that $\gT_g^z\vz_\star$ is the minimizer of $ \norm{\gT_g^x\vx_\star- G(\vz)}_2$. According to the assumption of the theorem, and as seen above,  $\hat{\vz}$ and $\vz_\star$ satisfy:
\[
\norm{\vy-\mA G(\hat{\vz})}_2\leq \norm{\vy-\mA\gT_g^x G(\vz_\star)}_2+\delta_{approx}.
\]
The goal is to convert this inequality to another one in terms of $\norm{\gT_g^x\vx_\star- G(\hat{\vz})}_2$ and $\norm{\vx_\star-G(\vz_\star)}_2$. The main steps are contained in the following derivation:
\begin{align}
    \norm{\gT_g^x G(\vz_\star)- G(\hat{\vz})}_2&=
    \norm{ G(\gT_g^z\vz_\star)- G(\hat{\vz})}_2 \tag{equivariance}
    \\
    &\leq \frac{\norm{\mA G(\gT_g^z\vz_\star)-\mA G(\hat{\vz})}_2+ \delta}{\gamma}\tag{$\srec$}\\
    &\leq \frac{ 
    \norm{\vy-\mA G(\hat{\vz})}_2 
    +\norm{\vy-\mA \gT_g^x G(\vz_\star)}_2 + \delta
    }{\gamma}
    \tag{triangle ineq.}\\
    &\leq \frac{ 
    2\norm{\vy-\mA \gT_g^x G(\vz_\star)}_2 +\delta_{approx}+ \delta
    }{\gamma}
    \tag{theorem assumption}\\
    &\leq \frac{ 
    2\norm{\mA\gT_g^x\vx_\star-\mA \gT_g^x G(\vz_\star)}_2 +2\norm{\vepsilon}_2+\delta_{approx}+ \delta
    }{\gamma}
    \tag{triangle ineq.}\\
    &\leq \frac{ 
    4\norm{\gT_g^x\vx_\star- \gT_g^x G(\vz_\star)}_2 +2\norm{\vepsilon}_2+\delta_{approx}+ \delta
    }{\gamma}\tag{condition (\ref{eq:norm_condition})}
\end{align}
These inequalities, Lemma 4.3 in \cite{Bora2017-as}, only require triangle inequalities, $\srec$  property and bound on the condition (\ref{eq:norm_condition}), where the last two conditions are satisfied by random Gaussian sensing matrices with i.i.d. entries and appropriately chosen variance. Using the following triangle inequality,
\begin{equation}
    \norm{\gT_g^x G(\vz_\star)- G(\hat{\vz})}_2\geq \norm{\gT_g^x\vx_\star- G(\hat{\vz})}_2 -\norm{\gT_g^x\vx_\star - \gT_g^x G(\vz_\star)}_2,
\end{equation}
and unitary property of $\gT_g^x$
\begin{equation}
\norm{\gT_g^x\vx_\star-\gT_g^x G(\vz_\star)}_2  = \norm{\vx_\star-  G(\vz_\star)}_2
\end{equation}
the theorem follows.

Let's consider the $\srec$  property and the condition (\ref{eq:norm_condition}). The later is satisfied by a random Gaussian matrix properly normalized. We use Gaussian concentration inequality to prove this.

\begin{theorem}[{Theorem 8.34  - \cite{foucart_mathematical_2013}}]
For a Lipschitz function $f:\R^n\to\R$ with Lipschitz constant $L$, if $\vg$ is a standard Gaussian random vector, then for all $t>0$, we have
\[
\mathbb{P}\left(f(\vg)-\E[f(\vg)]\geq t\right)\leq \exp\left(-\frac{t^2}{4L^2}\right).
\]
\end{theorem}
We choose $f$ as a function of the matrix $\mA$ and defined by $f(\mA)=\norm{\mA\vx}_2$. First, see that $f(\cdot)$ is Lipschitz with $L=\norm{\vx}_2$:
\[
\card{\norm{\mA\vx}_2 - \norm{\mB\vx}_2 }\leq \norm{\mA\vx-\mB\vx}_2\leq \norm{\mA-\mB}_2 \norm{\vx}_2\leq    \norm{\mA-\mB}_F \norm{\vx}_2.
\]
Therefore, we get, from the theorem, and for standard Gaussian random matrix $\mA$:
\[
\mathbb{P}(\norm{\mA\vx}_2-\E(\norm{\mA\vx}_2)\geq t)\leq \exp{\left(\frac{-t^2}{4\norm{\vx}^2_2}\right)}
\]
Note that $\E(f(\mA)=\E(\norm{\mA\vx}_2)$ can be upper bounded as: 
\[
\E(\norm{\mA\vx}_2)\leq (\E(\norm{\mA\vx}^2_2)^{1/2}=  (\sum_{i=1}^m \E(\inner{\va_i}{\vx})^2)^{1/2}  = (m\norm{\vx}^2_2)^{1/2}=\sqrt{m}\norm{\vx}_2.
\]
Finally, by choosing $t= \sqrt{m}\norm{\vx}_2$, we get with probability $1-e^{-m^2/4}$ that:
\[
\norm{\mA\vx}_2\leq t + \sqrt{m}\norm{\vx}_2 = 2 \sqrt{m}\norm{\vx}_2 
\]
Normalizing entries of the standard Gaussian random matrix $\mA$ by $1/\sqrt{m}$ will provide the desired result by choosing $\vx = \gT_g^x \vx^\star-\gT_g^x G(\vz_\star)$ which is a fixed vector independent of $\mA$.

The next step is, therefore, $\srec$ property. Since we use an equivariant generative model, the ambiguity about the rotation operation is captured in the latent space. Therefore, the very same approach of \cite{Bora2017-as} can be used to build $\epsilon$-nets and use that to establish $\srec$ for the equivariant generative model. We re-state this theorem for completeness here.

\begin{lemma}
For $L$-Lipshcitz equivariant generative model $G:\R^k\to\R^n$ and $\alpha<1$, the random Gaussian matrix $\mA\in\R^{m\times n}$ with i.i.d. entries $\gN(0,m^{-1})$ and $m =\Omega\left(\frac{k}{\alpha^2} \log (Lr/\delta)\right)$ satisfies $\srec(G(B_k(r)),1-\alpha,\delta)$ with probability $1-e^{-\Omega(\alpha^2m)}$.
\end{lemma}

Note that the proof of this lemma remains identical to \cite{Bora2017-as}, since the equivariance property does not play any role in the proof. It  is only necessary to establish it for a generative model $G$ and choosing $\gS=G(B_k(r))$.

As a final remark, consider for the moment the case of non-equivariant priors where the rotation needs to be extracted jointly with the latent code. In this case $\srec$ would translate to:
\begin{align}
    \norm{\gT_{g_\star}^x G(\vz_\star)- \gT_{\hat{g}}^x G(\hat{\vz})}_2&\leq 
 \frac{\norm{\mA \gT_{g_\star}^x G(\vz_\star)-\mA \gT_{\hat{g}}^x G(\hat{\vz})}_2+ \delta}{\gamma}.
\end{align}
In this case, we would need to establish $\srec$ for the set $\cup_{g\in\gG}\gT_{g}^x G(B_k(r))$. If the sets $\gT_{g}^x G(B_k(r))$ are disjoint for different $g$, the number of $\epsilon$-nets would increase, proportional to the $\epsilon$-net required to cover the group $\gG$. This would contribute another factor inside the logarithm for sample complexity.



\newpage
\section{Additional results} \label{appx:additional_exp}
\subsection{Eq-VAE: group choice}\label{appx:eq_vae_group}

\begin{figure}[h]
    \centering
    \begin{tabular}{cc}
        \includegraphics[width=0.4\textwidth]{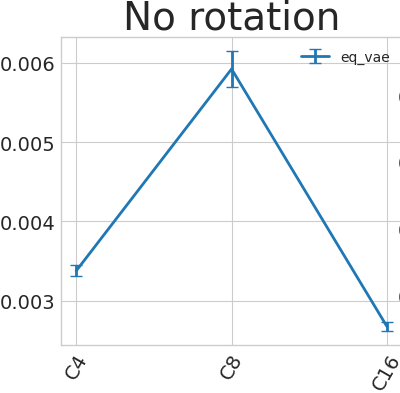} &
        \includegraphics[width=0.4\textwidth]{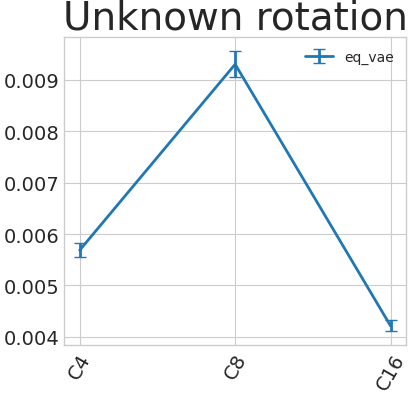} \\
    \end{tabular}
    \caption{Compressed sensing with 100 measurements on MNIST. We compare VAE equivariant under the cyclic groups $C_4$, $C_8$ and $C_{16}$.}
    \label{fig:vae_prior_group}
\end{figure}

Group choice can be rather important to the performance of compressed sensing. Thus, we compare results of three different prior: equivariant under the cyclic group of $4$, $8$ and $16$ rotations. In Figure \ref{fig:vae_prior_group} we report MSE for CS on MNIST with 100 measurements. All three priors perform well, but the best results are achieved when we use the cyclic group of 16 rotations. 

\newpage
\subsection{Eq-VAE: Samples}\label{appx:eq_vae_group_sample}
\begin{figure}[h]
    \centering
    \begin{tabular}{cc}
        \includegraphics[width=0.8\textwidth]{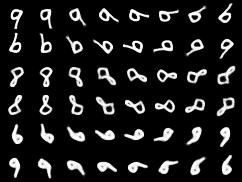} 
    \end{tabular}
    \caption{Samples from the equivariant VAE.}
    \label{fig:vae_prior_samples}
\end{figure}

We report the compressed sensing results for the $C_{16}$-equivariant VAE. Figure \ref{fig:vae_prior_samples} shows, how the reconstructions change when we apply the group action to a latent code and push it through the decoder (for three different latent codes). This way we observe that even though the model was not trained on the rotated images, it can generate them perfectly well.

\newpage
\subsection{Eq-VAE Priors for MAYO}\label{appx:eq_vae}

\begin{figure*}[ht!]
    \centering
    \includegraphics[width=0.5\textwidth]{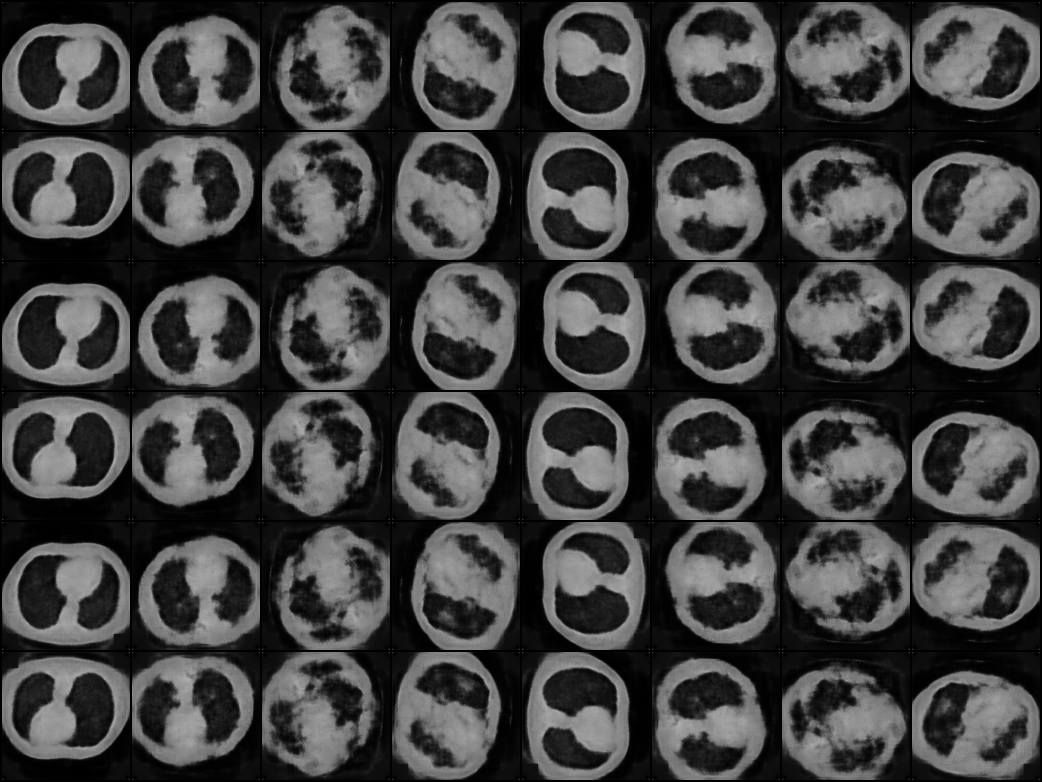}
    \caption{
    Rotated reconstructions from Eq-VAEs for MAYO dataset.}
    \label{fig:mayo_train_points}
\end{figure*}
\begin{figure*}[ht!]
    \includegraphics[width=\textwidth]{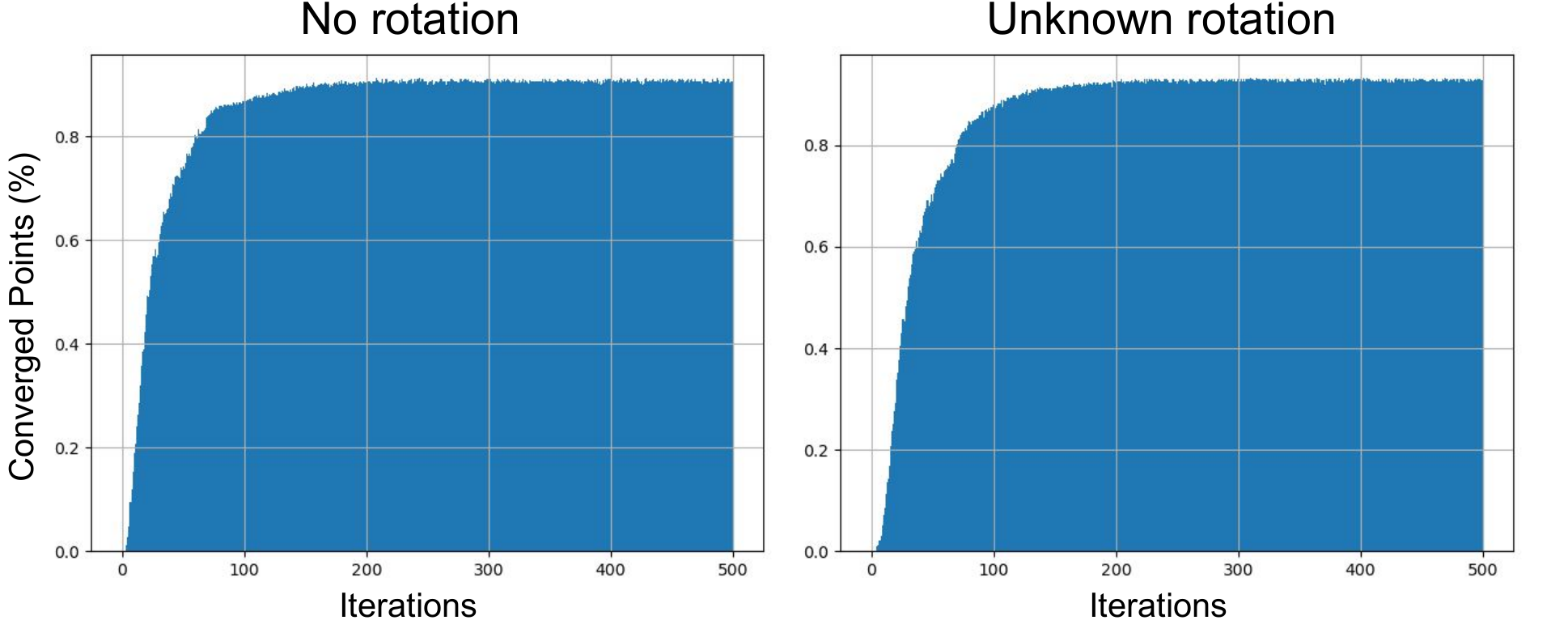}\hfill
    \caption{Number of iterations until convergence (i.e., MSE $\leq 1.e^{-2}$), for the MAYO dataset, using Eq-VAE as a generative prior. Left: No rotation angle. Right: Unknown rotations.}
    \label{fig:mayo_cdf_mse_conv}
\end{figure*}
\begin{figure*}[ht!]
    \includegraphics[width=\textwidth]{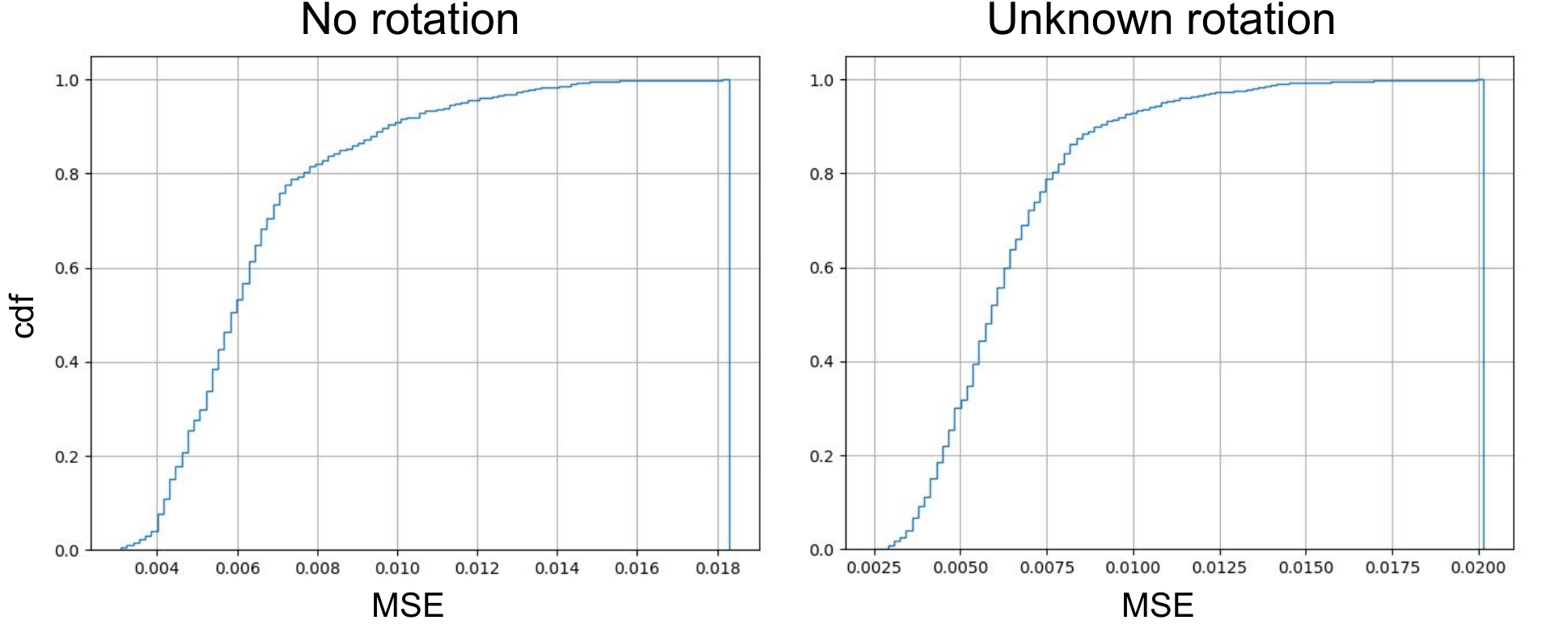}\hfill
    \caption{MSE from compressed sensing experiments, for the MAYO dataset, using Eq-VAE as a generative prior. Left: No rotation angle. Right: Unknown rotations.}
    \label{fig:mayo_cdf_mse_eqvae}
\end{figure*}
\begin{figure*}[ht!]
    \includegraphics[width=\textwidth]{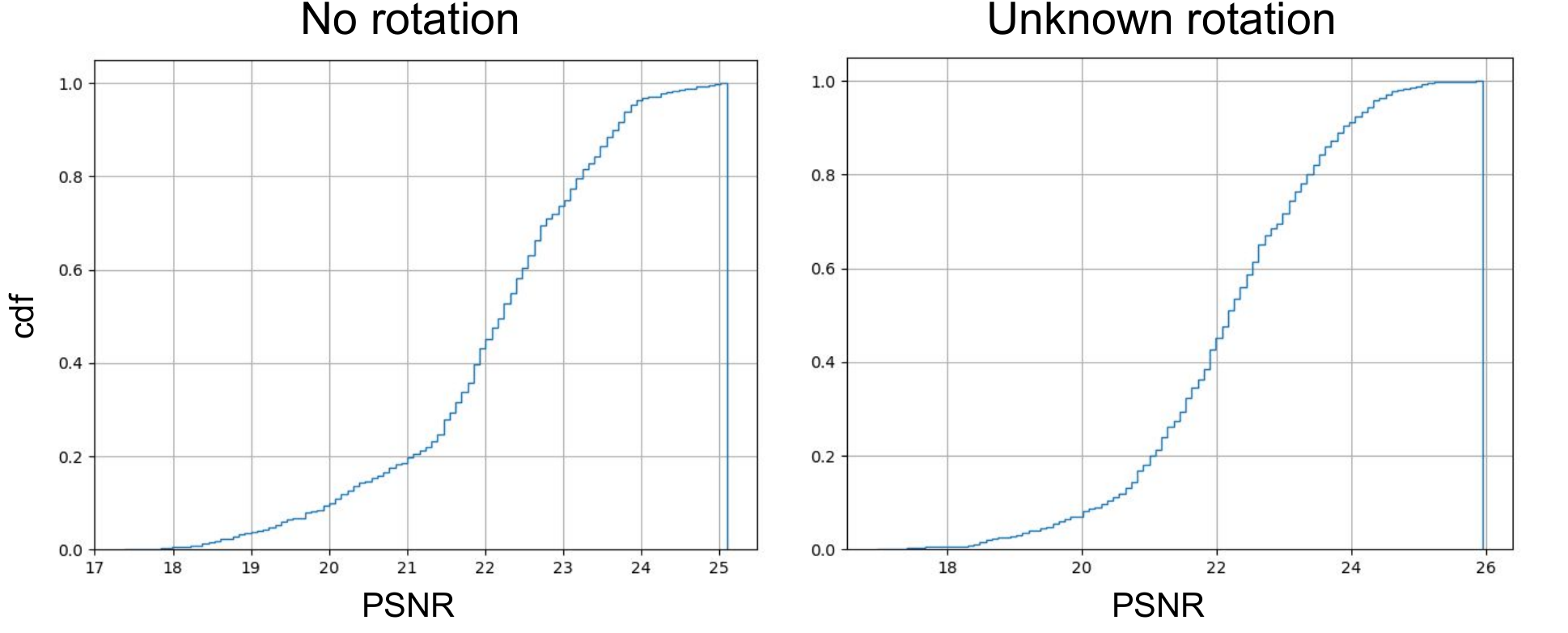}
    \caption{PSNR from compressed sensing experiments, for the MAYO dataset, using Eq-VAE as a generative prior. Left: No rotation angle. Right: Unknown rotations.}
    \label{fig:mayo_psnr}
\end{figure*}

\newpage
\section{Experimental Setup}\label{appx:exp_setup}

\subsection{Training Generative Prior}
\subsubsection{Convolutional VAE and Equivariant VAE}

\paragraph{Architecture} We train a VAE model, where both encoder and decoder are fully convolutional neural networks with the ReLU activations. 

The representation space size is set to $z_{dim} = 128$. 

For MNIST, the Conv-VAE encoder is fully convolutional architecture with kernel size, number of filters, stride, and padding as follows: $\text{Input signal (1)} \rightarrow [(3, 32, 2, 1) \rightarrow (3, 64, 2, 1) \rightarrow (3, 96, 2, 1) \rightarrow (3, 128, 2, 1) \rightarrow (3, 256, 2, 1) \rightarrow flatten(.) \rightarrow (\mu_z, \log\sigma_z^2)]$. The decoder comprises of transpose convolutional layers. Following the preceding format, the architectural details of decoder is as follows: $\text{Input signal (128)} \rightarrow [(3, 128, 1, 0) \rightarrow (3, 96, 1, 0) \rightarrow (3, 64, 1, 1) \rightarrow (4, 32, 2, 1) \rightarrow (4, 1, 2, 1) \rightarrow flatten(.) \rightarrow (\mu_x)]$.

For MAYO, the Conv-VAE encoder is fully convolutional architecture with kernel size, number of filters, stride, and padding as follows: $\text{Input signal (1)} \rightarrow [(5, 32, 1, 0) \rightarrow (3, 32, 1, 0) \rightarrow (3, 64, 2, 0) \rightarrow (3, 64, 1, 0) \rightarrow (3, 128, 2, 0) \rightarrow (3, 128, 2, 0) \rightarrow (3, 256, 2, 0) \rightarrow (3, 256, 2, 0) \rightarrow (2, 256, 2, 0) \rightarrow flatten(.) \rightarrow (\mu_z, \log\sigma_z^2)]$. The decoder comprises of transpose convolutional layers. Following the preceding format, the architectural details of decoder is as follows: $\text{Input signal (128)} \rightarrow [(6, 256, 1, 0) \rightarrow (3, 256, 2, 0) \rightarrow (3, 128, 2, 0) \rightarrow (4, 128, 1, 0) \rightarrow (3, 64, 2, 0) \rightarrow (3, 32, 2, 0) \rightarrow (4, 32, 1, 0) \rightarrow (3, 1, 1, 0) \rightarrow flatten(.) \rightarrow (\mu_x)]$.

For MAYO, the Equivariant VAE encoder comprises of group equivariant convolutional architecture with kernel size, number of filters, stride, and padding as follows: $\text{Input signal (1)} \rightarrow [(5, 16, 1, 0) \rightarrow (3, 16, 1, 0) \rightarrow (3, 16, 2, 0) \rightarrow (3, 16, 2, 0) \rightarrow (3, 32, 2, 0) \rightarrow (3, 32, 1, 0) \rightarrow (3, 32, 1, 0) \rightarrow (3, 64, 1, 0) \rightarrow (2, 256, 2, 0)  \rightarrow flatten(.) \rightarrow (\mu_z, \log\sigma_z^2)]$. The decoder comprises of equivariant transpose convolutional layers. Following the preceding format, the architectural details of decoder is as follows: $\text{Input signal (128)} \rightarrow [(6, 128, 1, 0) \rightarrow (3, 32, 2, 0) \rightarrow (3, 128, 2, 0) \rightarrow (4, 32, 1, 0) \rightarrow (3, 16, 2, 0) \rightarrow (3, 16, 2, 0) \rightarrow (4, 8, 1, 0) \rightarrow (3, 8, 1, 0) \rightarrow (3, 1, 1, 0) \rightarrow flatten(.) \rightarrow (\mu_x)]$.

Note: Aug-VAE and Aug-Flow follow the same architecture as Conv-VAE and Flow respectively. The only difference between augmented and their non-augmented counterpart is that the augmented models are trained on augmented (rotated) dataset.

\subsubsection{Multi-scale RealNVP}
\paragraph{Architecture}
The architecture of RealNVP comprises of 8 real-NVP blocks with scaling set to $3$.


\end{document}